\pdfoutput=1

\documentclass[11pt]{article}

\usepackage[]{acl}

\usepackage{times}
\usepackage{latexsym}
\usepackage{graphicx}

\usepackage{graphicx}
\usepackage{booktabs}
\usepackage{amsmath}

\usepackage[T1]{fontenc}

\usepackage[utf8]{inputenc}

\usepackage{microtype}

\usepackage{inconsolata}

\title{Frequency Explains the Inverse Correlation of Large Language Models' Size, Training Data Amount, and Surprisal's Fit to Reading Times}

\author{Byung-Doh Oh$^{\textnormal{1}}$ \qquad Shisen Yue$^{\textnormal{2}}$ \qquad William Schuler$^{\textnormal{1}}$ \\
  $^{\textnormal{1}}$The Ohio State University \qquad $^{\textnormal{2}}$Shanghai Jiao Tong University\\
  \texttt{oh.531@osu.edu} \qquad \texttt{lyw520@sjtu.edu.cn} \qquad \texttt{schuler.77@osu.edu}}

\begin{document}
\maketitle
\begin{abstract}
Recent studies have shown that as Transformer-based language models become larger and are trained on very large amounts of data, the fit of their surprisal estimates to naturalistic human reading times degrades.
The current work presents a series of analyses showing that word frequency is a key explanatory factor underlying these two trends.
First, residual errors from four language model families on four corpora show that the inverse correlation between model size and fit to reading times is the strongest on the subset of least frequent words, which is driven by excessively accurate predictions of larger model variants.
Additionally, training dynamics reveal that during later training steps, all model variants learn to predict rare words and that larger model variants do so more accurately, which explains the detrimental effect of both training data amount and model size on fit to reading times.
Finally, a feature attribution analysis demonstrates that larger model variants are able to accurately predict rare words based on both an effectively longer context window size as well as stronger local associations compared to smaller model variants.
Taken together, these results indicate that Transformer-based language models' surprisal estimates diverge from human-like expectations due to the superhumanly complex associations they learn for predicting rare words.
\end{abstract}

\section{Introduction}
The predictability of linguistic material in its context has been shown to be an important factor of real-time processing difficulty \citep{hale01, levy08}, with a large body of empirical work showing surprisal \citep{shannon48} to be a strong predictor of relevant behavioral and neural measures \citep[][\textit{i.a.}]{dembergkeller08, smithlevy13, haleetal18, shainetal20}.
Therefore, a core research goal of expectation-based theories of sentence processing has been to characterize the latent probability distribution of the human comprehender.
Language models (LMs) that define a conditional probability distribution are helpful for exploring these questions, as they can be trained to embody different predictive processes and yield concrete surprisal estimates that can be evaluated against measures of processing difficulty.

Recent work using surprisal estimates from Transformer-based LMs has revealed a strong inverse correlation between the size of LMs and the fit of their surprisal estimates to naturalistic human reading times, where larger models yield surprisal estimates that are less predictive of reading times \citep{ohetal22, ohschuler23tacl, shainetal22, devardamarelli23}.
Large amounts of training data have also been shown to play a detrimental role, with fit to reading times starting to degrade after LMs see about two billion tokens \citep{ohschuler23emnlp}.
This robust inverse correlation is meaningful, as it shows that increasingly larger LMs are less appropriate as models of human cognition, and that human sentence processing is not driven by the predictions LMs make with more model parameters and training data.
While open-class words like nouns and adjectives have been identified as driving the adverse effect of model size \citep{ohschuler23tacl}, how model size and the training data interact during LM training to give rise to such systematic divergence from human-like expectations remains unclear.

Studies on the scaling behavior of large LMs have recently shown that larger models learn examples faster by increasing their probabilities to a greater extent given the same amount of exposure \citep{tirumalaetal22}.
However, during early training stages, models of all sizes exhibit similar next-token predictions by learning to accurately predict frequent function words \citep{xiaetal23}.
This suggests that the difference in surprisal estimates as a function of model size will be modulated by frequency and will increase as models see larger amounts of training data.

Based on these observations, this work presents a series of analyses showing that word frequency is a key explanatory factor of the inverse correlation between model size, training data amount, and surprisal's fit to reading times.
First, residual errors from four LM families on four corpora show that the inverse correlation between model size and fit to reading times is the strongest on the subset of least frequent words, which is driven by excessively accurate predictions of larger model variants.
Moreover, training dynamics reveal that all model variants learn to predict rare words during later training steps and larger model variants do so more accurately, which explains the detrimental effect of both training data amount and model size on fit to reading times.
Finally, a feature attribution analysis demonstrates that larger model variants predict rare words more accurately compared to smaller model variants based on both an effectively longer context window and stronger local associations.
These results provide evidence that Transformer-based LMs' surprisal estimates diverge from human-like expectations due to the superhumanly complex associations they learn for predicting rare words.

\section{Experiment 1: Effect of Frequency on Strength of Inverse Correlation} \label{sec:exp1}

The first experiment examines the influence of word frequency on the strength of the inverse correlation between model size and fit to reading times by evaluating surprisal estimates from four LM families on four corpora of naturalistic reading times collected through both self-paced reading and eye-tracking paradigms.

\subsection{Response Data}
The reading times analyzed in this experiment come from the Natural Stories Corpus \citep{futrelletal21}, the Dundee Corpus \citep{kennedyetal03}, the Ghent Eye-Tracking Corpus \citep[GECO;][]{copetal17}, and the Provo Corpus \citep{lukechristianson18}.
The Natural Stories Corpus contains self-paced reading times from 181 subjects that read 10 naturalistic English stories consisting a total of 10,245 words.
The Dundee Corpus contains eye-gaze durations from 10 subjects that read 67 English newspaper editorials consisting a total of 51,501 words.
The GECO contains eye-gaze durations from 14 monolingual subjects that read the English version of the novel \textit{The Mysterious Affair at Styles} \citep{christie20}, which consists of 13 chapters and 56,441 words.
The Provo Corpus contains eye-gaze durations from 84 subjects that read 55 short English passages consisting a total of 2,746 words that range from news articles, science magazines, and works of fiction.

For the Natural Stories Corpus, data points were filtered to exclude those for sentence-initial and final words, those from subjects who answered fewer than four comprehension questions correctly, and those shorter than 100 ms or longer than 3000 ms, which resulted in 384,905 observations in the exploratory set.
For the three eye-tracking corpora, data points were filtered to remove those for unfixated words, words following saccades longer than four words, and words at starts and ends of sentences, screens, documents, and lines.
This resulted in a total of 98,115, 144,850, and 52,960 observations in the exploratory sets of the Dundee Corpus, GECO, and the Provo Corpus respectively.\footnote{The exploratory set of each corpus consists of roughly 50\% of all data points based on the sum of subject ID and sentence number. The held-out set is reserved for statistical significance testing and not analyzed in this work.}
All observations were log-transformed before regression modeling, following previous work \citep[e.g.][]{ohschuler23tacl}.\footnote{The log-transform implicitly assumes a superlinear linking function between surprisal and reading times, which has been shown to produce tighter fits for surprisal from larger LM variants \citep{shainetal22, hooveretal23}.}

\begin{figure*}[ht!]
    \includegraphics[width=\textwidth]{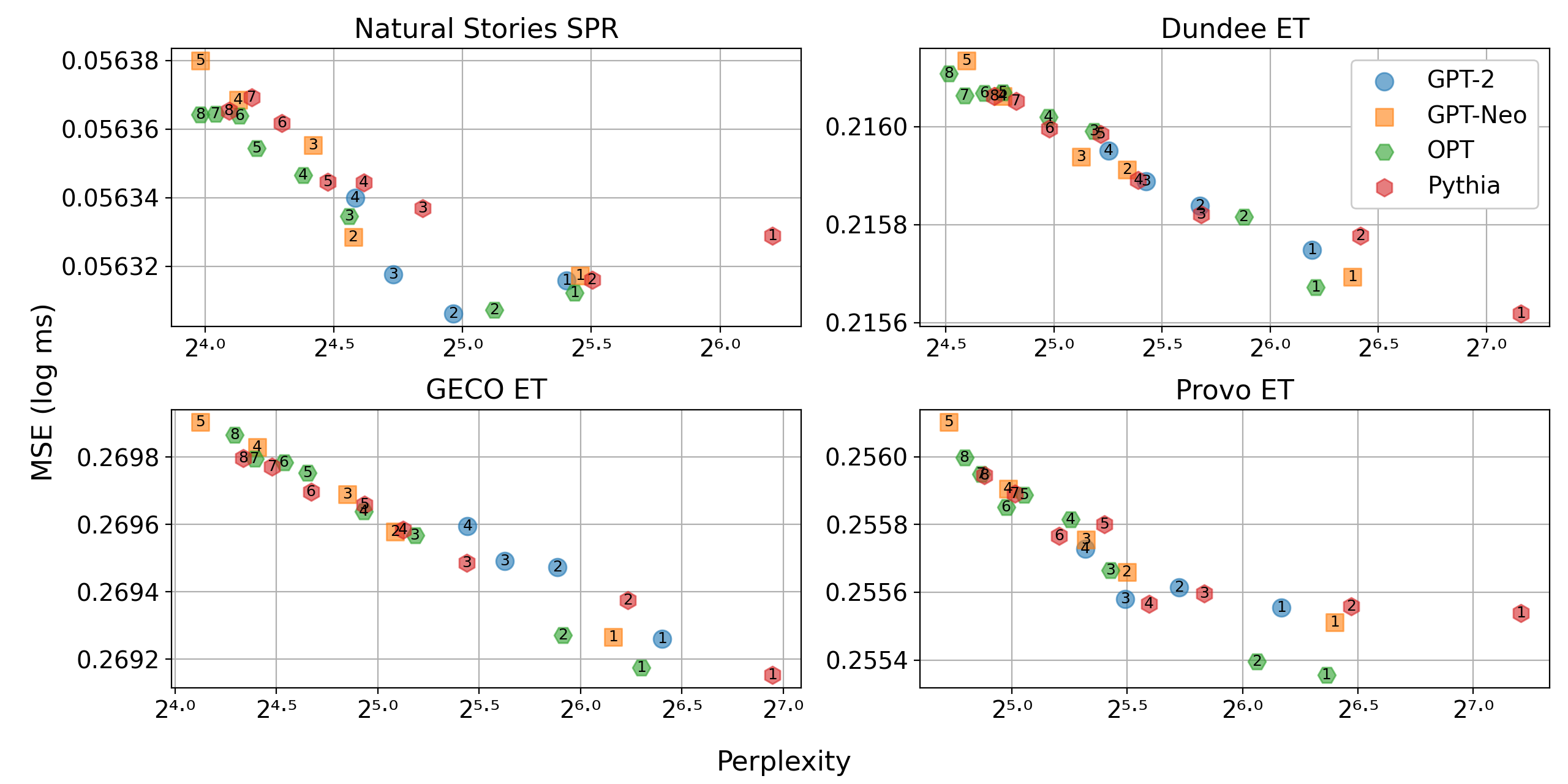}
\caption{Corpus-level perplexity measures from each GPT-2, GPT-Neo, OPT, and Pythia model variant, and mean squared errors of regression models that include each surprisal predictor on the four corpora of reading times. The ordered labels represent variants of different sizes, where `1' represents the smallest variant of each LM family.}
\label{fig:ppl_mse}
\end{figure*}

\subsection{Predictors} \label{sec:preds}
This experiment evaluates surprisal estimates from variants of four LM families, namely the GPT-2 \citep{radfordetal19}, GPT-Neo \citep{blacketal21,blacketal22,wangkomatsuzaki21}, OPT \citep{zhangetal22}, and Pythia \citep{bidermanetal23} families.
All of these LMs are autoregressive Transformer-based models whose variants differ primarily in their size.
The hyperparameters of all examined variants are outlined in Appendix \ref{app:variants}.

Each chapter or article of the four corpora was tokenized using each LM's respective byte-pair encoding \citep[BPE;][]{sennrichetal15} tokenizer and provided to all variants to calculate surprisal estimates, i.e.~$-\log_{2}\mathsf{P}( w_{i+1} \mid w_{1..i} )$.
In cases where each chapter or article did not fit completely into one context window, surprisal estimates for the remaining tokens were calculated by conditioning on the second half of the previous context window.

In addition to these surprisal predictors, a set of baseline predictors that capture low-level processing was also included in all regression models.
These predictors are word length in characters, index of word position within each sentence, unigram surprisal (both self-paced reading and eye-tracking corpora), as well as saccade length and whether the previous word was fixated (eye-tracking corpora only).
Unigram surprisal was estimated using counts of $\sim$33 billion pre-tokenized tokens from the Pile \citep{gaoetal20}, which is a collection of English language datasets.\footnote{Whenever a word consisted of multiple subword tokens, token-level unigram surprisal was summed to calculate the word-level unigram surprisal. Code for calculating LM surprisal and unigram surprisal is available at \url{https://github.com/byungdoh/llm_surprisal}.}
All predictors were standardized by centering and scaling before model fitting, and `spillover' versions of predictors were not included in the regression models to avoid convergence issues and simplify the analyses.

\subsection{Regression Modeling}
Subsequently, a set of linear mixed-effects (LME) models that contain one surprisal predictor and the baseline predictors outlined in Section \ref{sec:preds} were fit to self-paced reading times of the Natural Stories Corpus and go-past durations\footnote{Go-past durations were analyzed in this work as regressive eye movements are thought to reflect additional processing difficulty that is incurred by the current word.} of the three eye-tracking corpora using \texttt{lme4} \citep{batesetal15}.
All LME models included by-subject random slopes for all fixed effects and random intercepts for each subject.
Additionally, random intercepts were included for each subject-sentence interaction for self-paced reading times collected from 181 subjects, and random intercepts were included for each sentence for eye-gaze durations collected from smaller subject pools.
After the regression models were fit, their predictions were subtracted from the target reading times to calculate the residual errors for each regression model.

To examine the effect of word frequency on the strength of inverse correlation between model size and fit to reading times, the data points in each corpus were divided into quintiles according to unigram log-probabilities of the target word.
Subsequently, the mean squared errors (MSEs) from each regression model were calculated on each quintile.
The corpus-level perplexity of each model variant is also calculated and reported as a proxy for model size, based on a preliminary analysis showing very little difference in surprisal values among very large LM variants.

\begin{figure*}[ht!]
    \includegraphics[width=\textwidth]{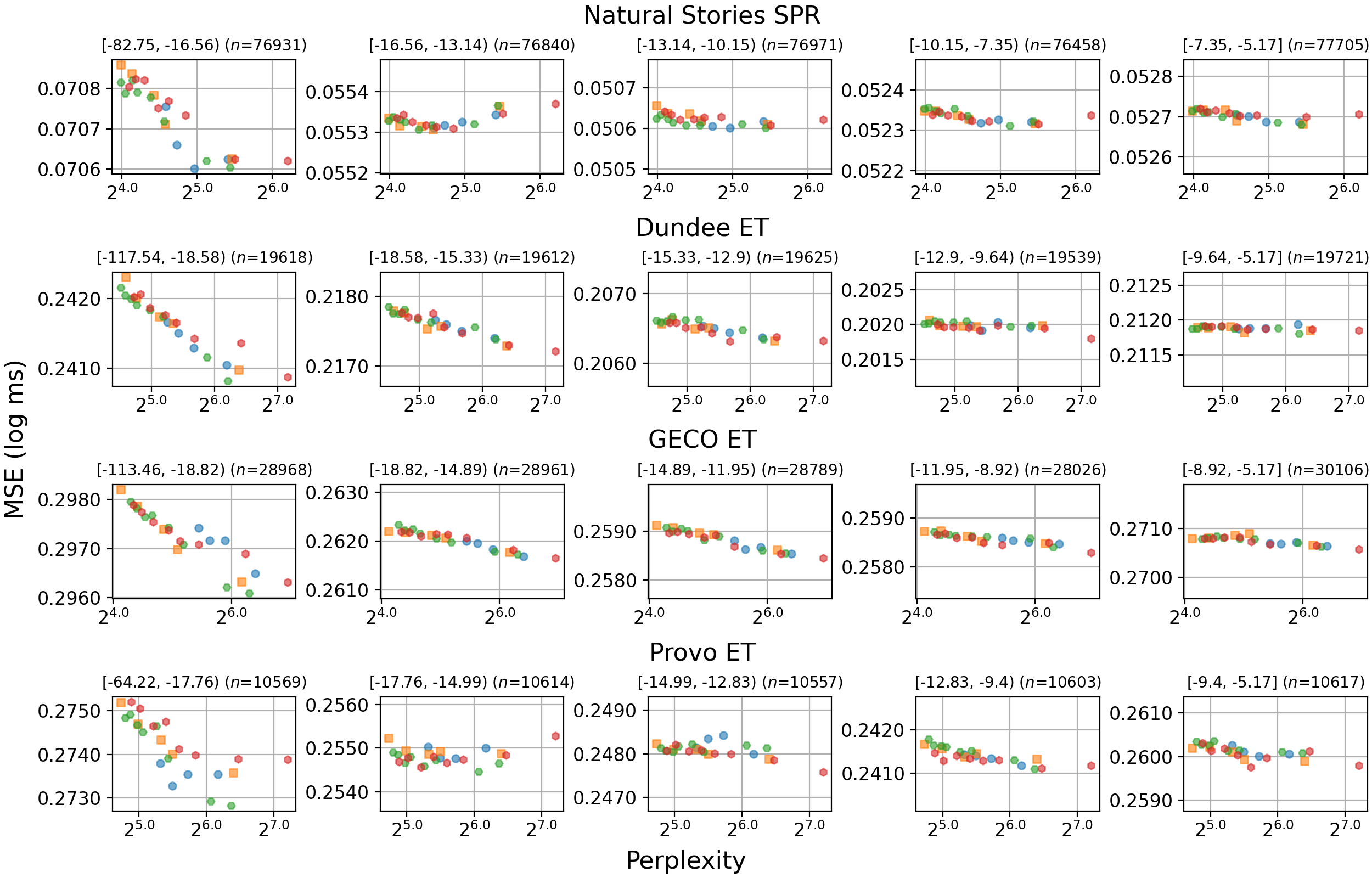}
\caption{Corpus-level perplexity measures from each GPT-2, GPT-Neo, OPT, and Pythia model variant, and mean squared errors of regression models that include each surprisal predictor on the four corpora of reading times. The columns of subplots represent the five quintiles defined by unigram log-probabilities.}
\label{fig:uniprob_mse}
\end{figure*}

\begin{figure*}[ht!]
    \includegraphics[width=\textwidth]{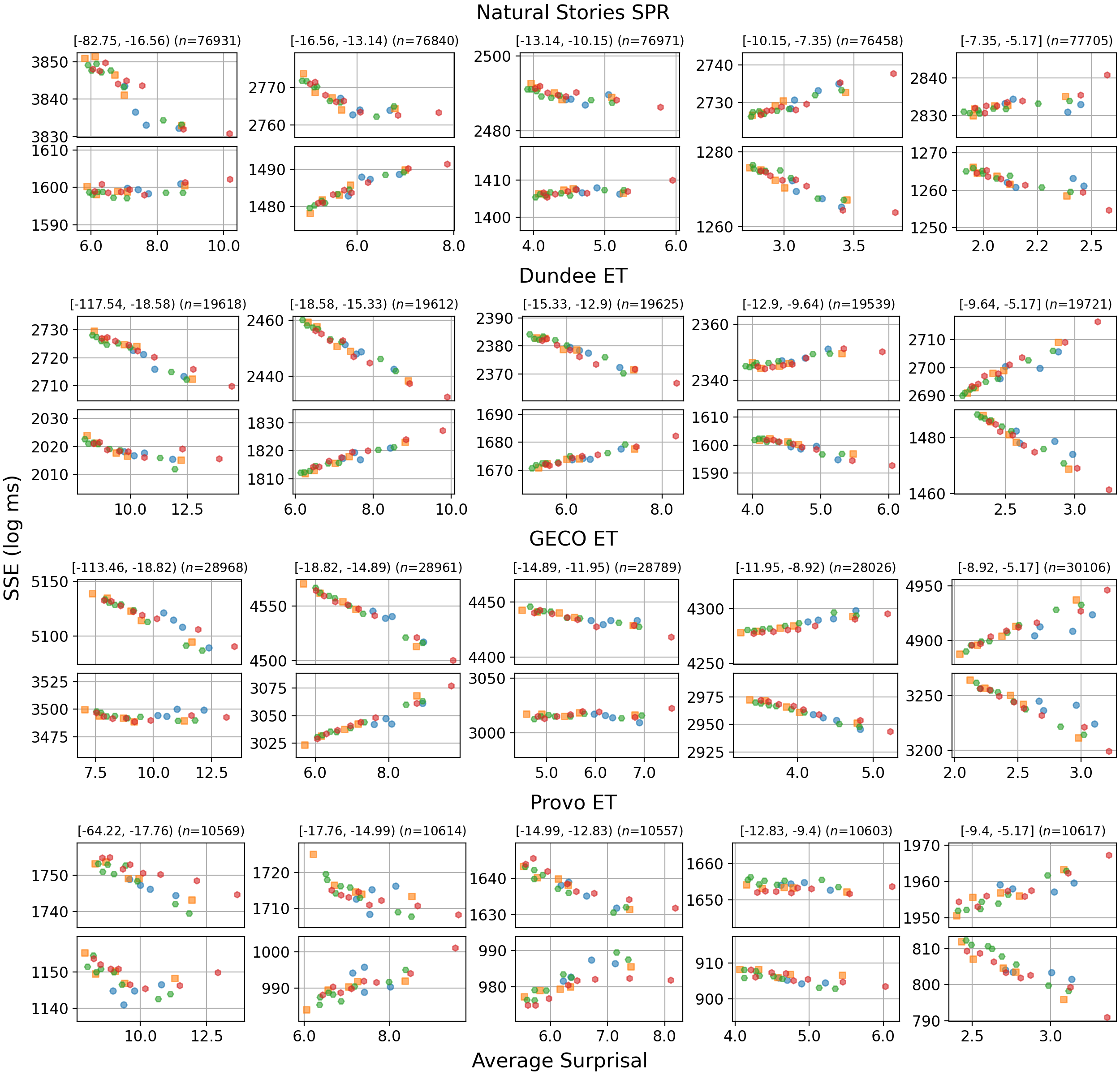}
\caption{Average surprisal values from each GPT-2, GPT-Neo, OPT, and Pythia model variant, and sum of squared errors of regression models that include each surprisal predictor on the four corpora of reading times. The columns of subplots represent the five quintiles defined by unigram log-probabilities. The top and bottom subplots of each row represent values from underpredicted and overpredicted data points, respectively.}
\label{fig:uniprob_sse}
\end{figure*}

\subsection{Results}
The results in Figure \ref{fig:ppl_mse} show that regression models with surprisal predictors from smaller model variants generally had lower MSEs across the four LM families on all four corpora.\footnote{The best-fitting lines between log perplexity and MSE had a slope significantly lower than 0 at $p < 0.05$ level by a one-tailed $t$-test on all four corpora.}
This replicates previous results reporting the inverse correlation between model size and fit to reading times \citep{ohschuler23tacl} and provides further empirical support for the effect.

The MSEs partitioned according to unigram probabilities in Figure \ref{fig:uniprob_mse} reveal that this corpus-level effect is primarily driven by the subset of least frequent words.
On all four corpora, the negative slope between log perplexity and MSE is the steepest on the first quintile, which is significantly lower than the other subsets at $p<0.05$ level by a permutation test that shuffles the quintile membership of the word.
The MSEs of regression models are also the largest on the first quintile on all four corpora, which suggests that the subset of rare words is where surprisal estimates from Transformer-based LMs diverge from human reading times the most.

A breakdown of the average surprisal values and sum of squared errors\footnote{SSEs are presented instead of MSEs as each regression model had different numbers of underpredicted and overpredicted data points, which can distort the MSEs and obscure the overall trend of mispredictions.} (SSEs) from underpredicted and overpredicted data points of each quintile in Figure \ref{fig:uniprob_sse} shows that larger model variants make especially accurate predictions on the subset of least frequent words compared to their smaller counterparts, where the difference in average surprisal is the biggest between model variants.
While more severe underpredictions of reading times are mostly responsible for the overall trend on this subset, the overpredictions do not appear to cancel out the trend, as is the case on other quintiles.

In contrast, more severe overpredictions of reading times seem to drive the overall trend  on the subsets of more frequent words.
This is likely due more to the estimated regression coefficients rather than the surprisal predictors, the difference in which between model variants is much smaller compared to other subsets.
As surprisal predictors from larger LM variants are smaller in magnitude, the regression models assign them higher coefficients to predict the same target reading times, which results in a systematic overprediction given surprisal predictors of similar magnitudes.

\begin{figure*}[ht!]
    \includegraphics[width=\textwidth]{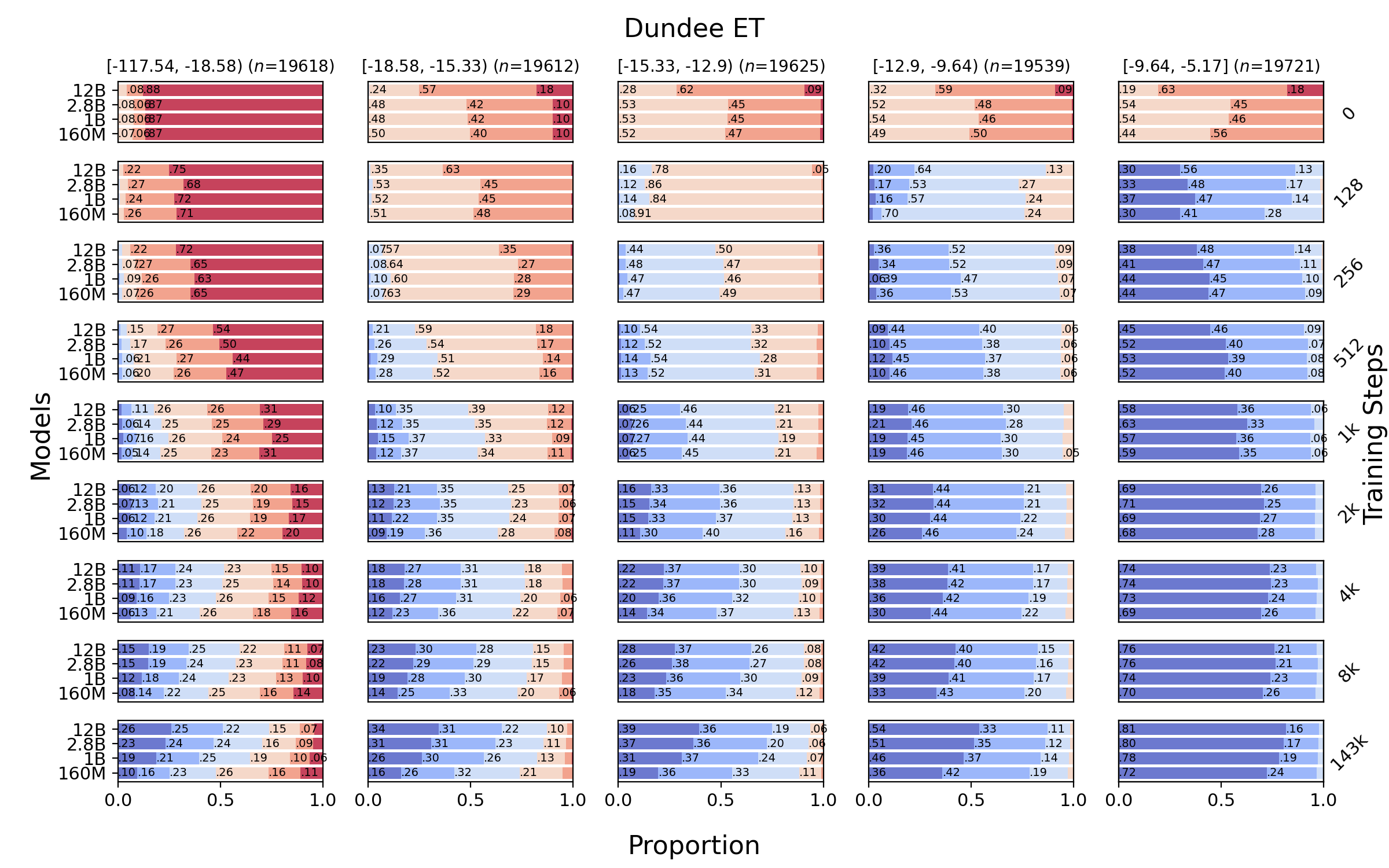}
    \includegraphics[width=\textwidth]{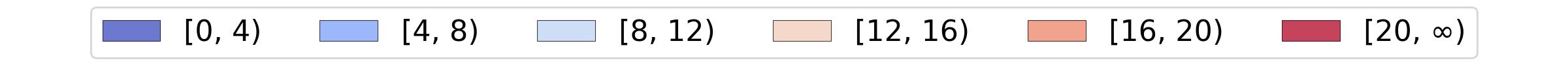}
\caption{Proportion of surprisal values from select Pythia model variants on each quintile of the Dundee Corpus as a function of training steps. Proportions that are greater than .05 are annotated.}
\label{fig:tsteps_surp}
\end{figure*}

\section{Experiment 2: Effect of Training Data Amount on Fit to Reading Times} \label{sec:exp2}
The previous experiment showed that the excessively accurate predictions of larger variants on the subset of least frequent words strongly drive the inverse correlation between model size and fit to human reading times.
The second experiment examines the training dynamics of Pythia model variants to study the influence of training data amount and model size on the ability of LMs to predict rare words, as well as the resulting fit of their surprisal estimates to human reading times.

\subsection{Procedures}
Among the four LM families examined in the previous experiment, the Pythia models are the only LMs that have publicly available checkpoints at various points during training.
These model variants were trained on batches of 1,024 examples with 2,048 tokens for a total of 143,000 training steps ($\sim$300 billion tokens).
Intermediate checkpoints that were saved during early training stages and after every 1,000 steps are publicly available.

To examine the training dynamics of these eight variants, surprisal estimates were calculated after \{0, 128, 256, 512, 1000, 2000, 4000, 8000, 143000\} training steps on the four corpora of reading times.\footnote{The Pythia variants evaluated in Experiment 1 are those that were fully trained for all 143,000 training steps. These steps were selected based on previous work that show a peak in fit to human reading times at around 1,000 training steps, and relatively little change after step 8,000 onwards for the Pythia models \citep{ohschuler23emnlp}.}
Subsequently, following identical regression modeling procedures as Experiment 1, LME models were fit to reading times and their residual errors were calculated.
Finally, the data points in each corpus were divided into quintiles according to unigram log-probabilities of the target word to examine the change in surprisal values and residual errors as a function of word frequency through the course of LM training.

\subsection{Results}
The surprisal values in Figure \ref{fig:tsteps_surp}\footnote{Data from select Pythia model variants on the Dundee Corpus is presented due to space constraints. Refer to Appendix \ref{app:surp} for comparable figures with all eight model variants on all four corpora.} show that at initialization (i.e.~after 0 training steps), surprisal values are the highest on the first quintile, as they are tokenized into multiple tokens by the BPE tokenizer.
During early training of up to 256 steps, all model variants primarily learn to predict frequent tokens, resulting in a large decrease in surprisal values on the fifth quintile.
As training continues, all model variants begin to learn to predict less frequent tokens, resulting in a consistent decrease in surprisal values on the lower quintiles.
Most notably, there seems to be no strong trend in surprisal values as a function of model size up until 1,000 training steps.

\begin{figure*}[ht!]
    \centering
    \includegraphics[width=\textwidth]{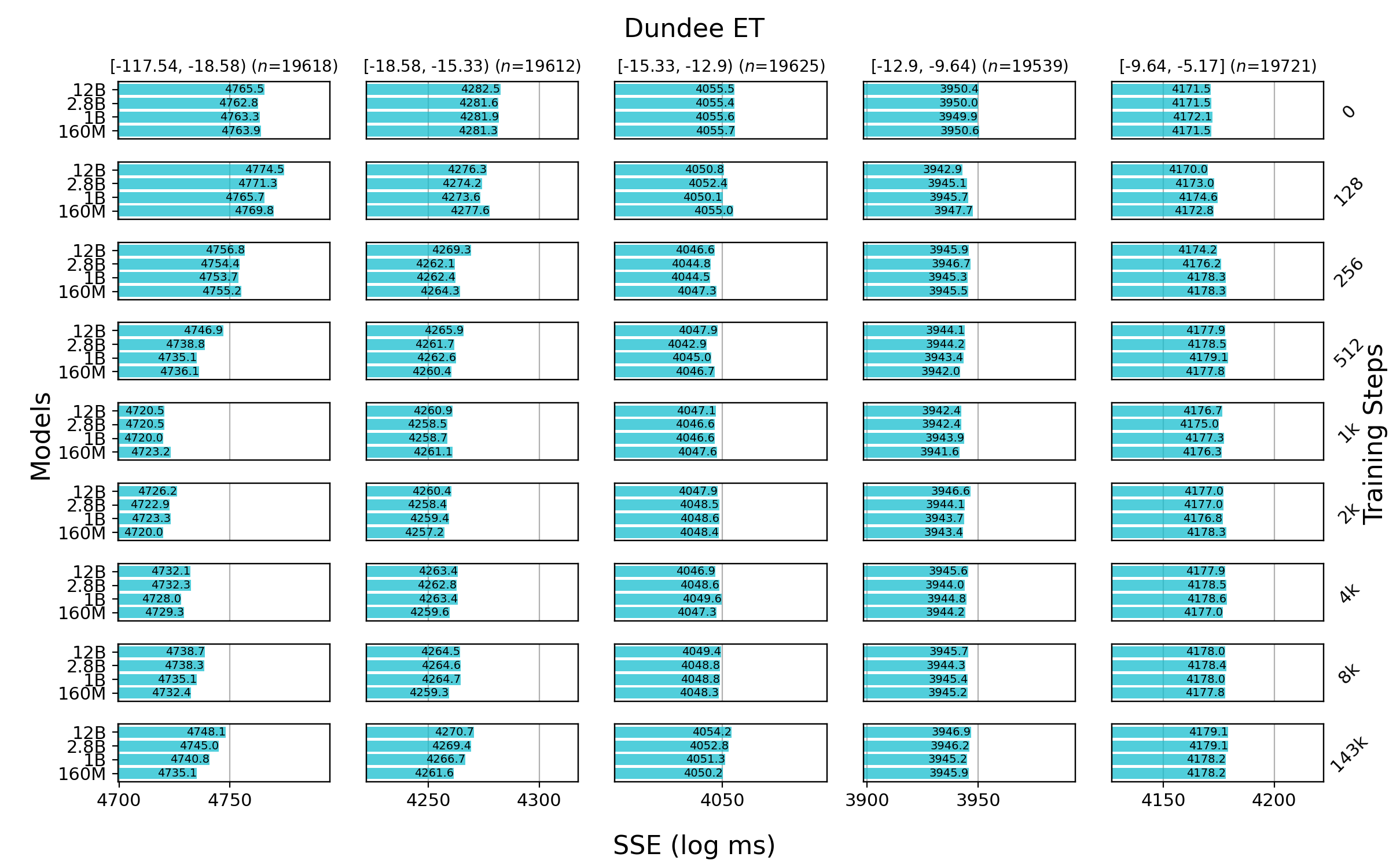}
\caption{SSEs from regression models containing surprisal predictors from select Pythia model variants on each quintile of the Dundee Corpus as a function of training steps. The columns of subplots share the scale but not the range of the x-axis for visual clarity.}
\label{fig:tsteps_sse}
\end{figure*}

In contrast, from training step 2,000 onward, the larger variants begin to yield lower surprisal values than their smaller counterparts, which suggests that the learning by larger variants actively continues while that by smaller variants begins to slow down.
By the end of training, this results in a general trend where the difference in surprisal values between model variants gets progressively bigger on the lower quintiles, as was demonstrated in Figure \ref{fig:uniprob_sse}.

The SSEs from regression models containing these surprisal predictors in Figure \ref{fig:tsteps_sse}\footnote{Refer to Appendix \ref{app:sse} for comparable figures with all eight model variants on all four corpora.} show that learning to predict rare tokens initially improves fit to reading times up to training step 1,000 by mostly improving the prediction of reading times of these words.
However, as the model variants see larger amounts of data and continue learning to predict rare tokens, the squared errors on these reading times begin to increase.
As larger model variants learn to do so more accurately, the increase in squared errors is also steeper for the corresponding regression models.
This demonstrates that both training data amount and model size help the accurate prediction of low-frequency words, which in turn has a detrimental effect on fit to reading times.

\begin{figure*}[ht!]
    \includegraphics[width=\textwidth]{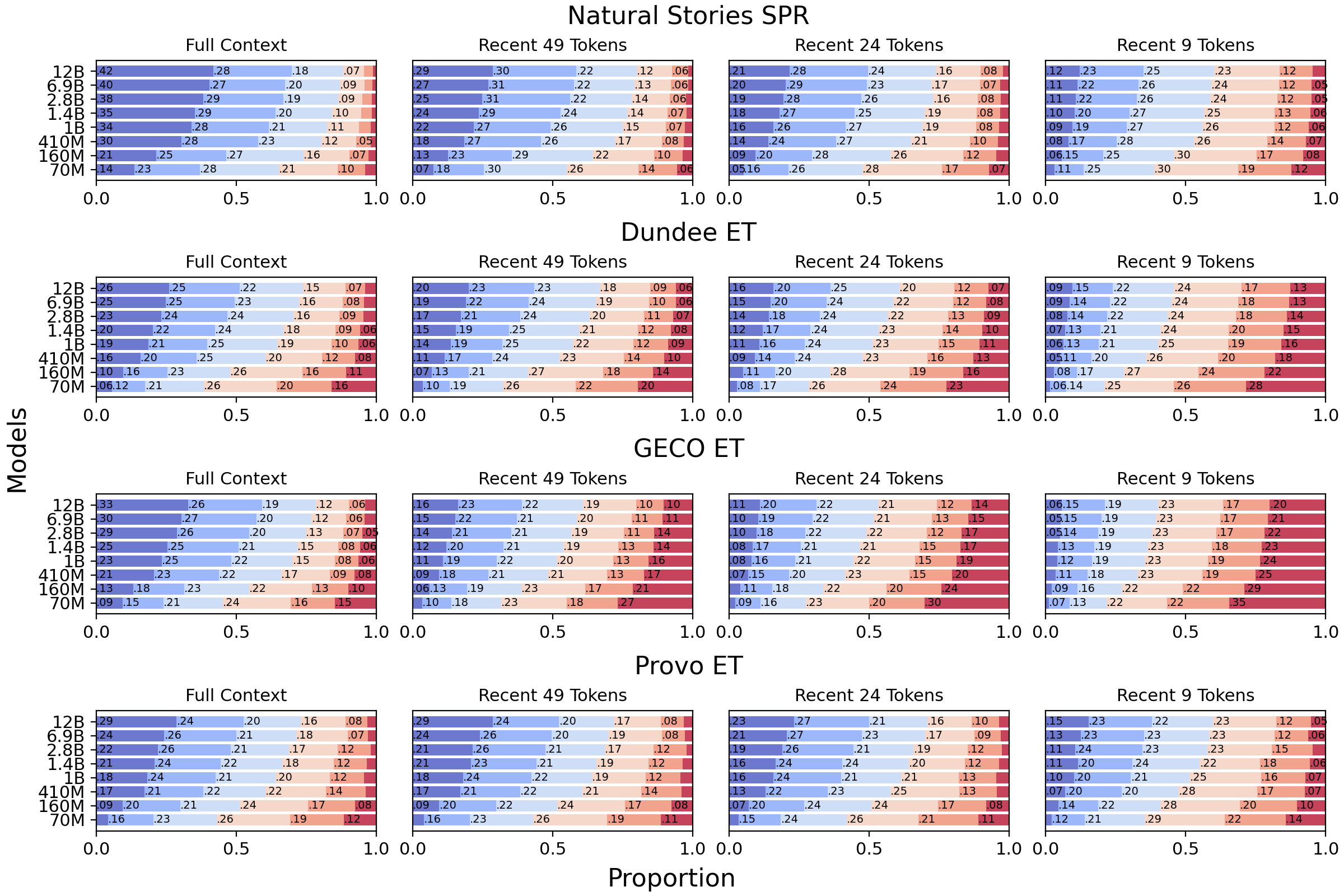}
    \includegraphics[width=\textwidth]{figures/legend_surp.png}
\caption{Proportion of surprisal values from Pythia model variants on the quintile of least frequent words of the four corpora of reading times as a function of context window size. Proportions that are greater than .05 are annotated.}
\label{fig:recentsurp}
\end{figure*}

\section{Follow-up Analysis: What Enables Larger Models to Predict Rare Words?} \label{sec:exp3}
The previous experiment showed that as LM variants are trained on large amounts of data, all variants learn to predict rare words during later training steps.
However, larger variants are able to do so more accurately, thereby ultimately resulting in surprisal estimates that are poorer predictors of human reading times.
The goal of this follow-up analysis is to provide a mechanistic explanation underlying the behavior of larger LM variants identified through the main experiments, by elucidating how the increase in model size enables LMs to make more accurate predictions of rare words.

Although the predictions of Transformer-based language models are opaque and difficult to interpret, one possible explanation for this phenomenon is that the larger variants have a longer `effective' context window size than smaller variants.
In other words, although all Transformers theoretically have veridical access to the context, the smaller variants may not be able to learn associations to early material in the context due to their limited capacity.
This follow-up analysis gauges the contribution of early and recent context tokens using a feature attribution method to examine what enables larger model variants to predict rare words more accurately than their smaller counterparts.

\subsection{Procedures}
The feature attribution method adopted for this analysis is based on limiting the LM's context to the most recent $n$ tokens \citep{kuribayashietal22}, which can also be viewed as an ablation of the earlier context tokens \citep[\textit{occlusion};][]{zeilerfergus14}.
This method has the advantages that it quantifies the contribution of context tokens in terms of interpretable probabilities (cf.~gradient-based methods) and does not suffer from potential out-of-distribution issues \citep{hookeretal19} as it keeps contiguous $n$-grams intact and does not arbitrarily alter their vector representations.

Surprisal estimates from the eight fully-trained Pythia variants were calculated on the quintile of least frequent words of the four reading time corpora by conditioning on the most recent \{49, 24, 9\} context tokens.
The resulting surprisal values were compared to those calculated by conditioning on the full context to examine the impact of early and recent context tokens.
If the early material in context is crucial for a model variant to predict rare tokens, larger increases in their surprisal values will be observed when the context window is limited.

\subsection{Results}
The results in Figure \ref{fig:recentsurp} show that on all four corpora, surprisal values from all variants progressively increase as the context window becomes more limited.
This increase seems to be modulated by model size, where larger variants demonstrate larger degrees of increase as a result of limiting the context window.
This suggests that larger variants have learned stronger associations with early context tokens and therefore have effectively longer context windows for predicting rare tokens.

However, surprisal values calculated under the most restrictive nine-token condition show that larger model variants are still able to make more accurate predictions than their smaller counterparts even when the context is very limited.
Taken together, these results indicate that larger model variants are able to predict rare words more accurately based on both an effectively longer context window as well as stronger local associations.

\section{Discussion} \label{sec:discussion}
This work presents word frequency as a unified explanation for the degradation in fit of surprisal estimates to naturalistic human reading times as a function of LM size and training data amount observed in recent studies.
First, evaluation of model variants from four LM families on four corpora of reading times shows that the inverse correlation between model size and fit to reading times is the strongest on the subset of least frequent words.
This is due to the larger model variants' disproportionately accurate predictions on this subset, where the target reading times are generally longer.
These findings are consistent with those from the analyses of \citet{ohschuler23tacl}, who found that more severe underpredictions of reading times of open-class words like nouns and adjectives most strongly drive the trend between model size and regression model fit.
Word frequency provides a more general and parsimonious account of the trend on the entire corpus, as well as a complementary view of the phenomenon.

The training dynamics of LMs also show a strong interaction between frequency and model size through the course of training.
More specifically, model variants of all sizes learn to accurately predict frequent tokens and also show little difference in surprisal values during early training steps.
However, as they continue to see large amounts of data, they start learning to predict rare tokens accurately.
It is at this later point in training where the difference between model variants begins to manifest, with larger variants learning to make more accurate predictions of these rare tokens.
These trends are consistent with prior work on the scaling behavior of large LMs \citep{tirumalaetal22, xiaetal23}, as well as observations that neural LMs first approximate unigram and then bigram probabilities during early training \citep{changbergen22,changetal23}.

Residual errors from regression models containing surprisal estimates at intermediate points during LM training show that learning to predict rare tokens initially improves fit to reading times by primarily improving the prediction of reading times of these rare words.
Nonetheless, as the model variants continue learning to predict rare tokens, the errors on these reading times begin to increase.
Since larger model variants learn to predict rare tokens more accurately, the regression models containing their surprisal estimates also exhibit a steeper increase in errors.
This illustrates the detrimental effect of training data amount and model size on fit to reading times, and also explains \citeauthor{ohschuler23emnlp}'s \citeyearpar{ohschuler23emnlp} observation of the peak in fit to reading times at around two billion training tokens.

The follow-up feature attribution analysis that ablates the contribution of early context tokens suggests that larger model variants utilize both an effectively longer context window and stronger local associations to predict rare tokens more accurately than their smaller counterparts.
Limiting the number of tokens in the context window weakens these associations for predicting rare words, which is most likely the reason why this improves the fit of LM surprisal to reading times, as demonstrated by \citet{kuribayashietal22}.

Taken together, these results indicate that both model size and large amounts of training data allow Transformer-based LMs to learn superhumanly complex associations for predicting rare words, which in turn adversely affects fit to human reading times.
In other words, surprisal from model variants that are smaller and trained on less data yield a better fit to naturalistic reading times because they implicitly capture word frequency.
This has important implications for research into whether frequency effects are dissociable from predictability effects in naturalistic reading \citep[e.g.][]{goodkindbicknell21,shain19,shain23}.
One possible interpretation of the current results is that they provide support for a strong and dissociable frequency effect, as the subset of rare words is where LM surprisal estimates diverge most from naturalistic reading times as a whole.
However, they may also indicate that the excessive number of parameters and training data result in surprisal estimates that have washed out frequency effects which could have been explained by predictability.

An interesting direction for future work is extending the current analyses to data collected in other languages \citep[e.g.][]{kuribayashietal21, devardamarelli23, wilcoxetal23}.
Based on the training dynamics of LMs observed in this work, to the extent that they are of sufficient sizes and are trained on large amounts of data, the explanation based on word frequency is expected to robustly generalize to data from other languages.

\section{Conclusion} \label{sec:conclusion}
This work proposes word frequency as an explanation for the inverse correlation observed between Transformer-based LMs' size, training data amount, and surprisal's fit to human reading times.
Four LM families on four corpora show the strongest inverse correlation between model size and fit to reading times on the least frequent words, which is driven by the more accurate predictions of the larger variants.
Training dynamics reveal that all variants learn to predict rare words with large amounts of data and larger variants do so more accurately, which explains the detrimental effect of both model size and training data amount.
These results indicate that the superhumanly complex associations for predicting rare words make Transformer-based LMs' surprisal estimates diverge from human-like expectations.

\section*{Acknowledgments}
We thank the ARR reviewers and the area chair for their helpful comments.
This work was supported by the National Science Foundation grant \#1816891.
All views expressed are those of the authors and do not necessarily reflect the views of the National Science Foundation.

\section*{Limitations}
The explanation for the dissociation between surprisal estimates from Transformer-based language models and real-time comprehension difficulty developed in this work is based on language model variants trained on English text and data from subjects that are native speakers of English.
Therefore, the proposed explanation may not generalize to other languages.
Other possible limitations include the assumption of linear effects and the lack of spillover predictors in regression modeling.

\section*{Ethics Statement}
This work used data collected as part of previously published research \citep{futrelletal21, kennedyetal03, copetal17, lukechristianson18}.
Readers are referred to the respective publications for more information on the data collection and validation procedures.
As this work focuses on studying the connection between conditional probabilities of language models and human sentence processing, its potential negative impacts on society appear to be minimal.

\bibliography{custom}

\appendix

\section{Hyperparameters of LM Variants}
The hyperparameters of model variants from the four LM families evaluated in this work are outlined in Table \ref{tab:params}.
\label{app:variants}
\begin{table}[ht!]
    \centering
    \small
    \begin{tabular}{lrrrr} \toprule
    Model Variant & \#L & \#H & $d_{\text{model}}$ & \#Parameters \\ \hline
    GPT-2 Small & 12 & 12 & 768 & $\sim$124M \\
    GPT-2 Medium & 24 & 16 & 1024 & $\sim$355M \\
    GPT-2 Large & 36 & 20 & 1280 & $\sim$774M \\
    GPT-2 XL & 48 & 25 & 1600 & $\sim$1.6B \\ \hline
    GPT-Neo 125M & 12 & 12 & 768 & $\sim$125M \\
    GPT-Neo 1.3B & 24 & 16 & 2048 & $\sim$1.3B \\
    GPT-Neo 2.7B & 32 & 20 & 2560 & $\sim$2.7B \\
    GPT-J 6B & 28 & 16 & 4096 & $\sim$6B \\
    GPT-NeoX 20B & 44 & 64 & 6144 & $\sim$20B \\ \hline
    OPT 125M & 12 & 12 & 768 & $\sim$125M \\
    OPT 350M & 24 & 16 & 1024 & $\sim$350M \\
    OPT 1.3B & 24 & 32 & 2048 & $\sim$1.3B \\
    OPT 2.7B & 32 & 32 & 2560 & $\sim$2.7B \\
    OPT 6.7B & 32 & 32 & 4096 & $\sim$6.7B \\
    OPT 13B & 40 & 40 & 5120 & $\sim$13B \\
    OPT 30B & 48 & 56 & 7168 & $\sim$30B \\
    OPT 66B & 64 & 72 & 9216 & $\sim$66B \\ \hline
    Pythia 70M & 6 & 8 & 512 & $\sim$70M \\
    Pythia 160M & 12 & 12 & 768 & $\sim$160M \\
    Pythia 410M & 24 & 16 & 1024 & $\sim$410M \\
    Pythia 1B & 16 & 8 & 2048 & $\sim$1B \\
    Pythia 1.4B & 24 & 16 & 2048 & $\sim$1.4B \\
    Pythia 2.8B & 32 & 32 & 2560 & $\sim$2.8B \\
    Pythia 6.9B & 32 & 32 & 4096 & $\sim$6.9B \\
    Pythia 12B & 36 & 40 & 5120 & $\sim$12B \\ \bottomrule
    \end{tabular}
    \caption{Hyperparameters of model variants whose surprisal estimates were examined in this work. \#L, \#H, and $d_{\text{model}}$ respectively refer to number of layers, number of attention heads per layer, and embedding size.}
    \label{tab:params}
\end{table}

\section{Surprisal Values as a Function of Training Steps}
\label{app:surp}
\begin{figure*}[ht!]
    \includegraphics[width=\textwidth]{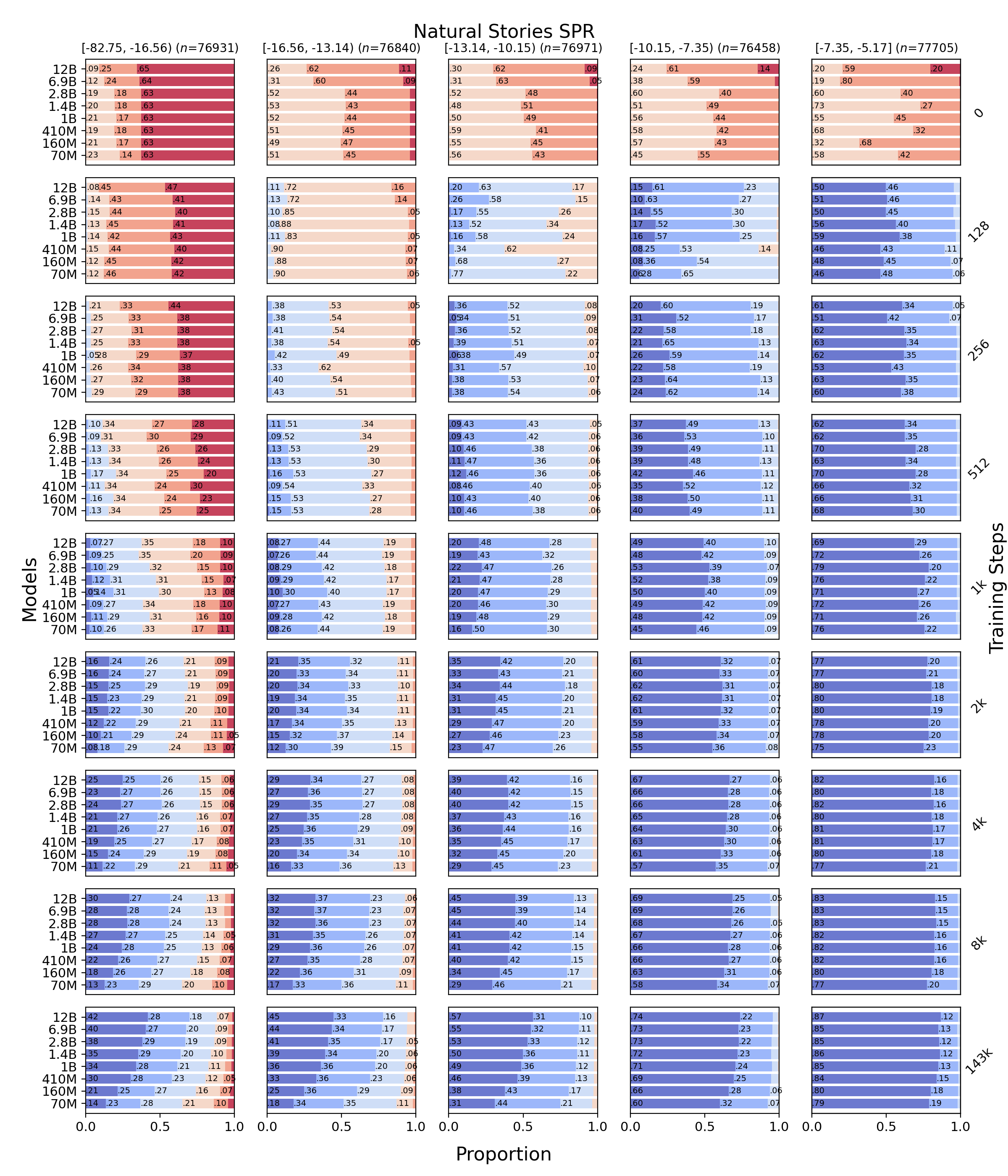}
    \includegraphics[width=\textwidth]{figures/legend_surp.png}
\caption{Proportion of surprisal values from Pythia model variants on each quintile of the Natural Stories Corpus as a function of training steps. Proportions that are greater than .05 are annotated.}
\label{fig:tsteps_surp_ns}
\end{figure*}

\begin{figure*}[ht!]
    \includegraphics[width=\textwidth]{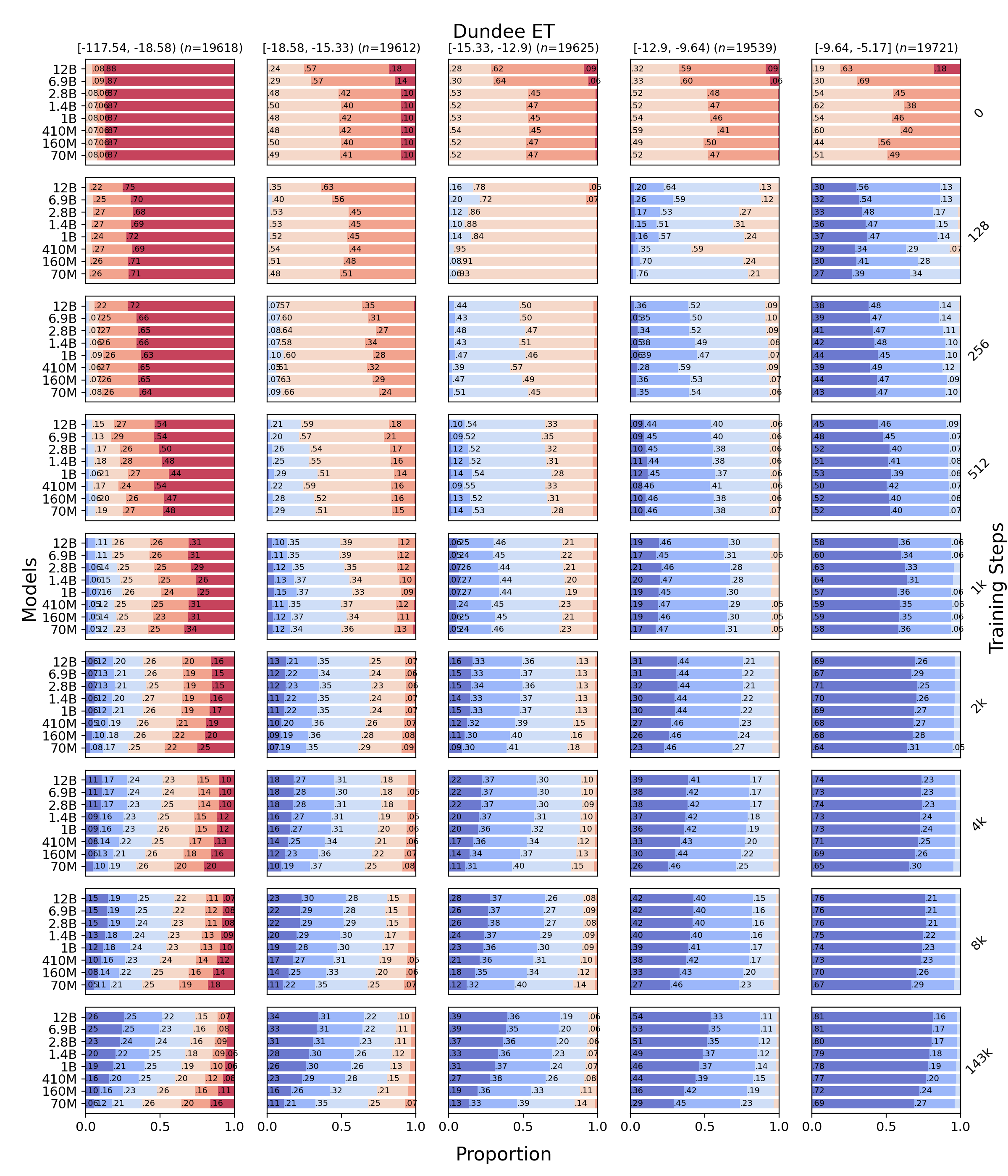}
    \includegraphics[width=\textwidth]{figures/legend_surp.png}
\caption{Proportion of surprisal values from Pythia model variants on each quintile of the Dundee Corpus as a function of training steps. Proportions that are greater than .05 are annotated.}
\label{fig:tsteps_surp_dd}
\end{figure*}

\begin{figure*}[ht!]
    \includegraphics[width=\textwidth]{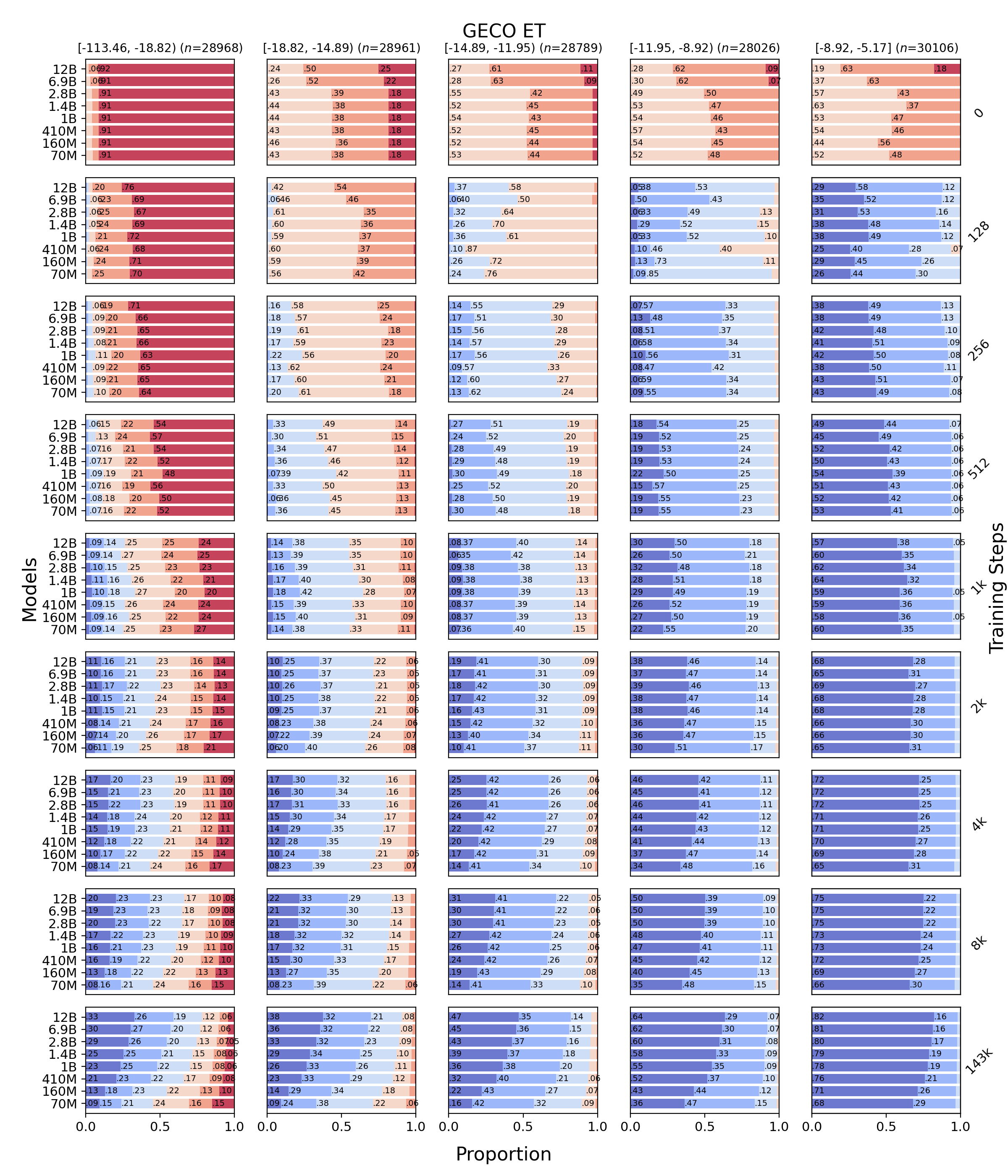}
    \includegraphics[width=\textwidth]{figures/legend_surp.png}
\caption{Proportion of surprisal values from Pythia model variants on each quintile of the GECO as a function of training steps. Proportions that are greater than .05 are annotated.}
\label{fig:tsteps_surp_gc}
\end{figure*}

\begin{figure*}[ht!]
    \includegraphics[width=\textwidth]{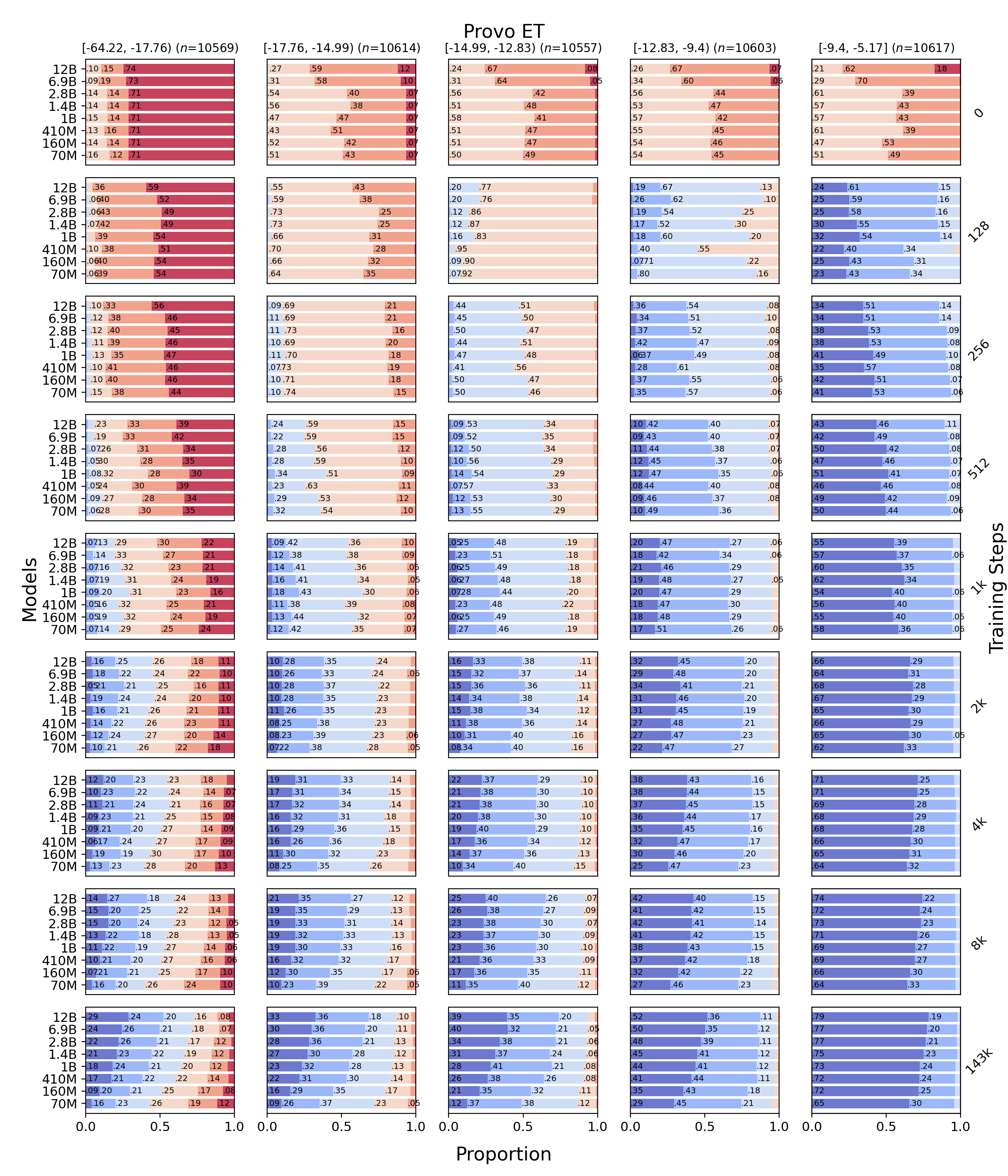}
    \includegraphics[width=\textwidth]{figures/legend_surp.png}
\caption{Proportion of surprisal values from Pythia model variants on each quintile of the Provo Corpus as a function of training steps. Proportions that are greater than .05 are annotated.}
\label{fig:tsteps_surp_pv}
\end{figure*}

The proportion of surprisal values from the Pythia LM family as a function of training steps on each quintile of the four corpora can be found in Figures \ref{fig:tsteps_surp_ns} (Natural Stories), \ref{fig:tsteps_surp_dd} (Dundee), \ref{fig:tsteps_surp_gc} (GECO), and \ref{fig:tsteps_surp_pv} (Provo).

\section{SSEs From Regression Models as a Function of Training Steps}
\label{app:sse}
\begin{figure*}[ht!]
    \centering
    \includegraphics[width=\textwidth]{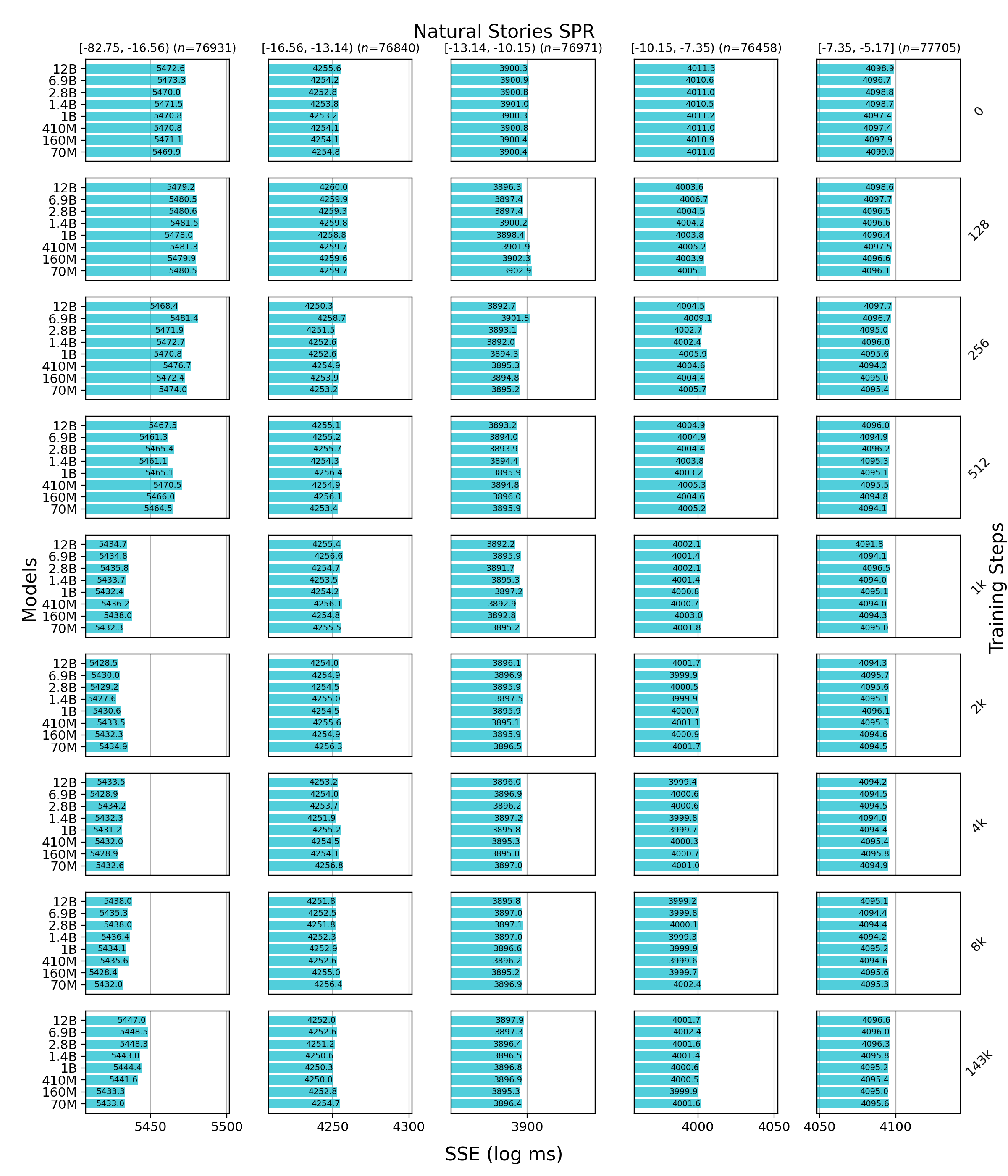}
\caption{SSEs from regression models containing surprisal predictors from Pythia model variants on each quintile of the Natural Stories Corpus as a function of training steps. The columns of subplots share the scale but not the range of the x-axis for visual clarity.}
\label{fig:tsteps_sse_ns}
\end{figure*}

\begin{figure*}[ht!]
    \centering
    \includegraphics[width=\textwidth]{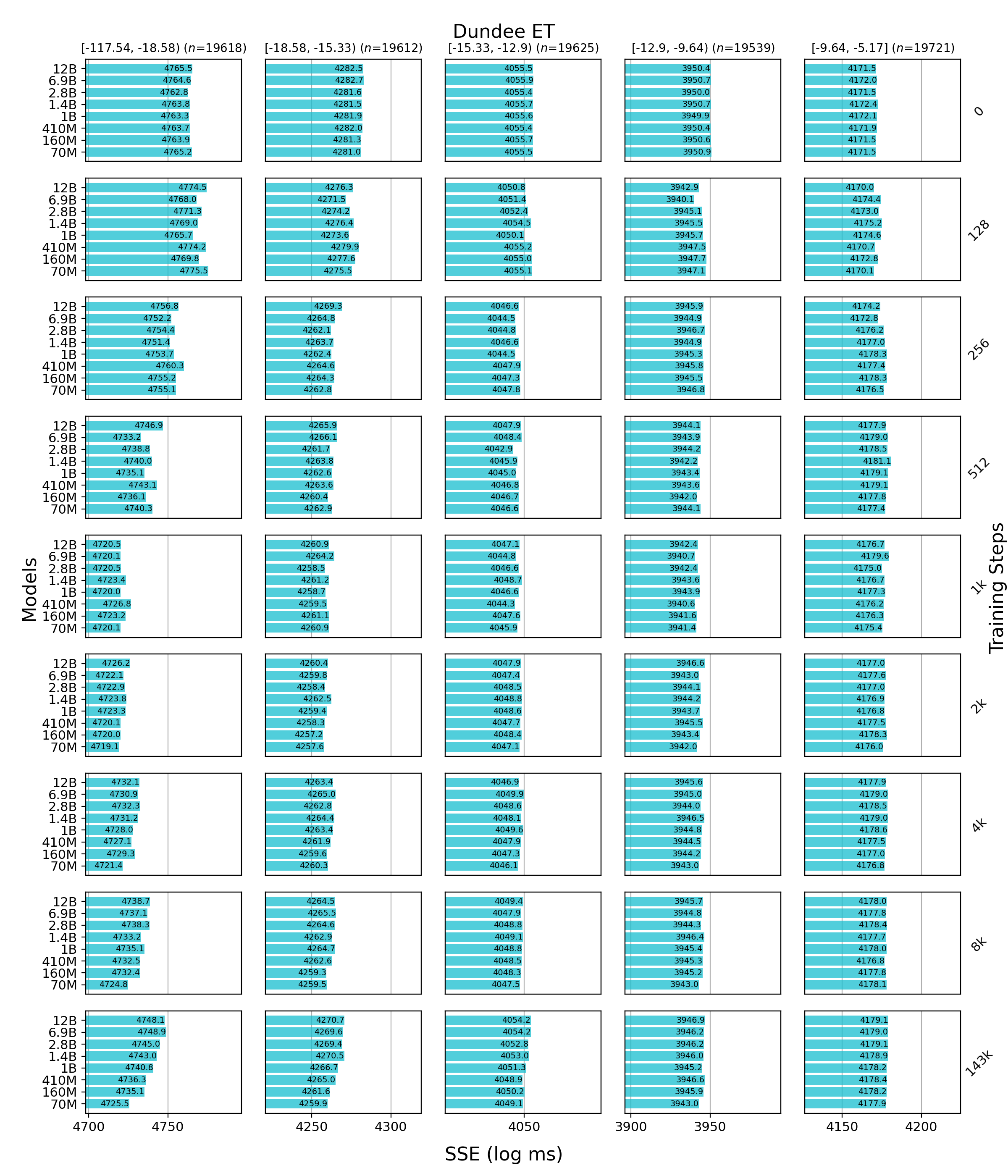}
\caption{SSEs from regression models containing surprisal predictors from Pythia model variants on each quintile of the Dundee Corpus as a function of training steps. The columns of subplots share the scale but not the range of the x-axis for visual clarity.}
\label{fig:tsteps_sse_dd}
\end{figure*}

\begin{figure*}[ht!]
    \centering
    \includegraphics[width=\textwidth]{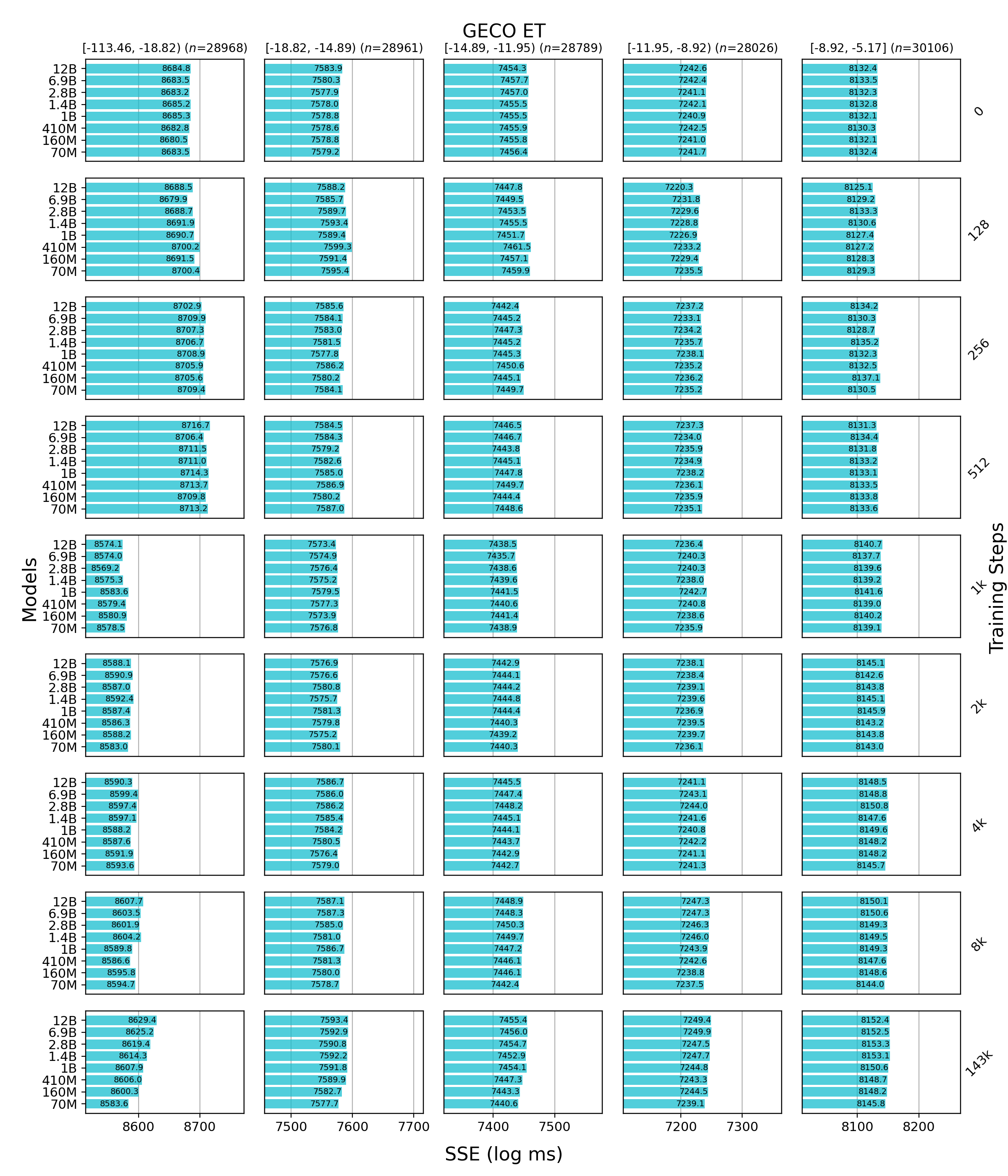}
\caption{SSEs from regression models containing surprisal predictors from Pythia model variants on each quintile of the GECO as a function of training steps. The columns of subplots share the scale but not the range of the x-axis for visual clarity.}
\label{fig:tsteps_sse_gc}
\end{figure*}

\begin{figure*}[ht!]
    \centering
    \includegraphics[width=\textwidth]{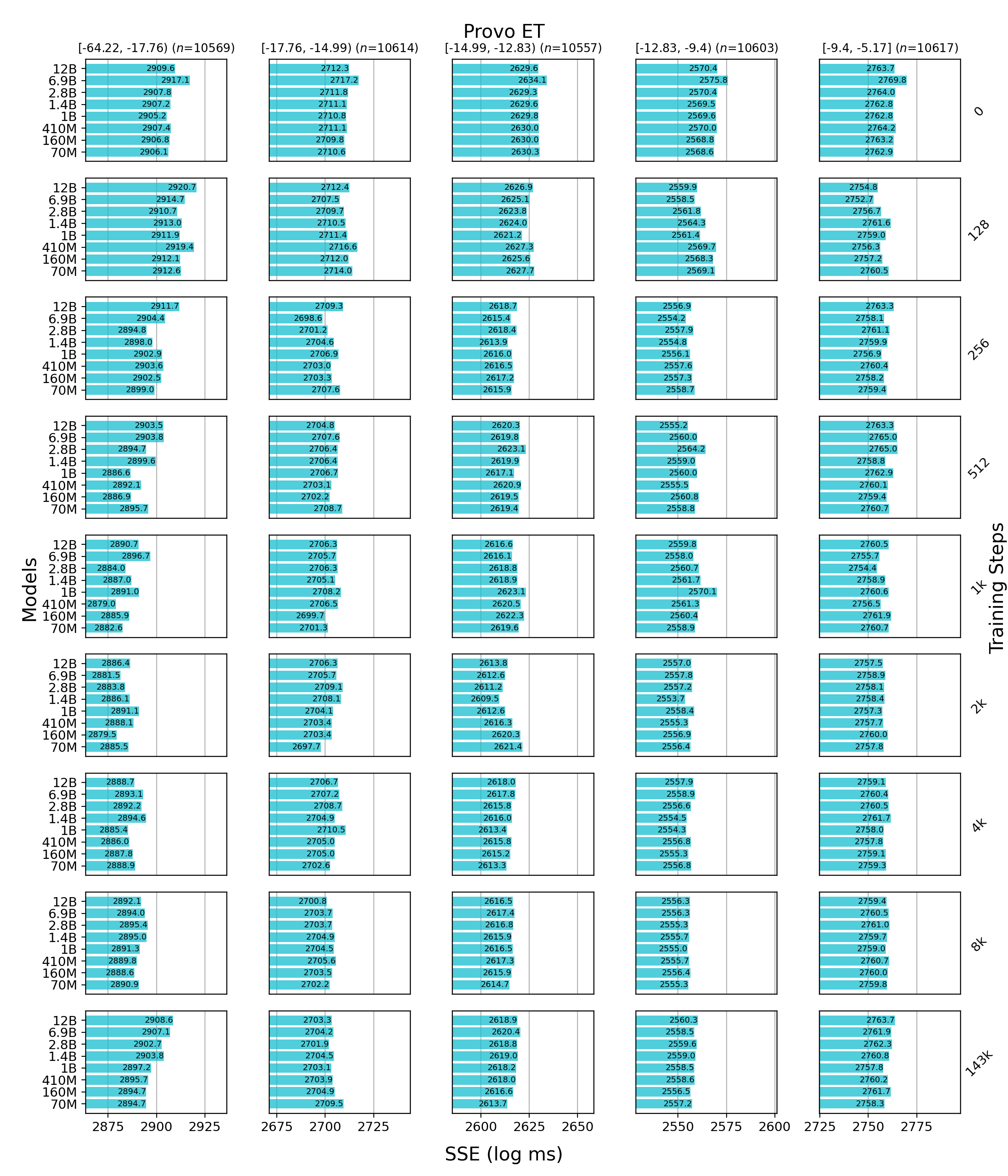}
\caption{SSEs from regression models containing surprisal predictors from Pythia model variants on each quintile of the Provo Corpus as a function of training steps. The columns of subplots share the scale but not the range of the x-axis for visual clarity.}
\label{fig:tsteps_sse_pv}
\end{figure*}

The SSEs from regression models containing surprisal predictors from the Pythia LM family as a function of training steps on each quintile of the four corpora can be found in Figures \ref{fig:tsteps_sse_ns} (Natural Stories), \ref{fig:tsteps_sse_dd} (Dundee), \ref{fig:tsteps_sse_gc} (GECO), and \ref{fig:tsteps_sse_pv} (Provo).

\end{document}